\documentclass{article} 
\usepackage{iclr2027_conference,times}

\usepackage{amsmath,amsfonts,bm}

\def\eqref#1{equation~\ref{#1}}

\def\1{\bm{1}}

\DeclareMathAlphabet{\mathsfit}{\encodingdefault}{\sfdefault}{m}{sl}
\SetMathAlphabet{\mathsfit}{bold}{\encodingdefault}{\sfdefault}{bx}{n}

\usepackage{hyperref}
\hypersetup{pdftitle={Distilling Text LMs into Speech-to-Speech LMs by Matching Internal Text Predictions},pdfauthor={Ju-Chieh Chou, Jiawei Zhou, Karen Livescu}}
\usepackage{url}
\usepackage{pifont}
\usepackage{booktabs}
\usepackage{graphicx}
\usepackage{tabularx}
\usepackage{wrapfig}
\usepackage{array}
\usepackage{float}
\usepackage[table]{xcolor}
\usepackage{colortbl}

\title{Distilling Text LMs into Speech-to-Speech LMs by Matching Internal Text Predictions}

\author{Ju-Chieh Chou\textsuperscript{1}, Jiawei Zhou\textsuperscript{2}, and Karen Livescu\textsuperscript{1}\\
\textsuperscript{1} TTI-Chicago \qquad \textsuperscript{2} Stony Brook University}
\date{}
\iclrfinalcopy

\begin{document}

\maketitle
\fancyhead{}

\begin{abstract}

Text language models (LMs) acquire broad linguistic and factual capabilities through large-scale pretraining, while comparable capabilities remain difficult to learn from speech alone. Recent work has shown that knowledge distillation from pretrained text LMs offers a promising way to transfer their linguistic and factual capabilities to speech models. However, existing distillation approaches have largely focused on speech-input, text-output models.  This idea has yet to be explored for {\it speech-to-speech} models, which operate on both speech input and speech output. In this work we introduce a speech-to-speech pretraining method that creates an internal text interface for knowledge transfer while preserving the speech-only input and output.
The model first consumes speech context without its transcription, switches to text prediction, and then uses the predicted text to lead future speech generation word by word. 
It also provides an interface for knowledge distillation: the internal text predictions can be directly distilled from the base text LM. To the best of our knowledge, this is the first application of text LM distillation to a speech-to-speech LM.
Under matched training conditions, the proposed sequence construction substantially improves the generative perplexity of spoken continuations over speech-only and interleaved speech-text models, approaching the performance of a text-leading inner-monologue model with external ASR prompt transcription. More importantly, distillation improves factual performance under both text and speech evaluation, indicating better retention of knowledge from the base text language model. 


\end{abstract}
\section{Introduction}
\begin{figure}[b]
    \centering
    \includegraphics[width=\linewidth]{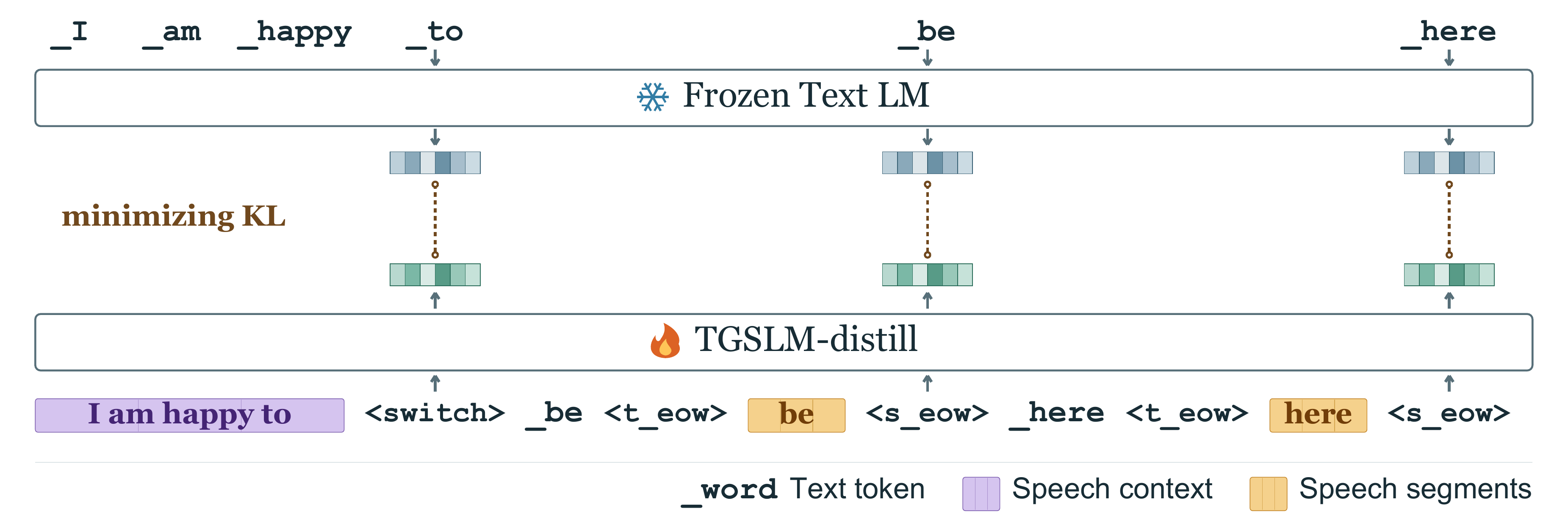}
    \caption{Our TGSLM architecture.
Given a speech context, the model generates subsequent words sequentially, with each text word followed by its corresponding speech segments. In TGSLM-distill, the text-token predictions are additionally trained to match those of a frozen text LM.}
    \label{fig:arch}
\end{figure}
Large-scale text pretraining provides the foundation for acquiring broad linguistic and factual capabilities. A text language model can first learn from massive unlabeled text corpora and subsequently transfer these capabilities to downstream settings such as instruction following, dialogue, and agents~\citep{ouyang2022training}. An analogous goal for speech language models (SLMs) is to acquire such capabilities directly from large-scale speech. However, scaling analyses suggest that comparable linguistic capabilities are harder to obtain from speech-only pretraining~\citep{cuervo2024scaling,maimon2025scaling}. In particular, under matched training compute budgets, performance on linguistic evaluations improves more slowly with scale for speech than for text. This motivates making more direct use of text during speech pretraining, both as an explicit training signal and as an intermediate prediction target for knowledge distillation from pretrained text language models.

\begin{wraptable}{r}{0.48\textwidth}   
  \vspace{-8pt}                       
  \centering
  \small
  \begin{tabular}{lcc}
    \toprule
    Model & Speech-only & Text-guided \\
          & prompt      & generation  \\
    \midrule
    Text leading & \ding{55} & \ding{51} \\
    Interleaved     & \ding{51} & \ding{55} \\
    TGSLM           & \ding{51} & \ding{51} \\
    \bottomrule
  \end{tabular}
  \caption{How text is used at inference in SLMs.}
  \label{tab:tradeoff}
\end{wraptable}
One way to use text is to incorporate it directly into the speech language modeling objective. Existing SLMs do so in different ways, leading to different tradeoffs. Interleaved models~\citep{nguyen2025spirit,chou2023toward} train on sequences that alternate between speech and text spans
from the same utterance, typically switching modalities at word boundaries, and can perform speech-to-speech inference without any text input. In this setting, text provides an additional training signal, but speech generation does not condition on text, resulting in worse linguistic performance~\citep{chou2026quantifying}. Text-leading, or inner-monologue, models~\citep{defossez2024moshi} instead generate text ahead of speech so text predictions directly condition subsequent speech generation. In prompted continuation, however, this typically requires the transcript of the speech prompt at inference to maintain the text-leading stream. Table \ref{tab:tradeoff} summarizes this tradeoff. This raises a question: how can we use text to guide linguistic prediction in SLMs without requiring prompt text at inference?

We introduce Text-Guided Speech LM (TGSLM), which approaches this question by separating future speech prediction into its linguistic content and its acoustic realization, as shown in the bottom part of Fig.~\ref{fig:arch}. Rather than predicting future speech directly from the speech context, we decompose generation into first predicting the future linguistic content in text and then generating the corresponding speech conditioned on the predicted text. During training, we insert a switch token at a randomly selected word boundary in the speech sequence. Before the switch, the model consumes speech directly. After the switch, it generates the continuation word by word, first predicting each text word and then generating its corresponding spoken realization. At inference, the switch token explicitly triggers the transition to text-guided speech generation. This construction captures complementary advantages of interleaved and text-leading models: text directly guides future speech generation, while inference requires no transcription of the speech prompt.

The TGSLM prediction structure substantially improves generative perplexity (genPPL). TGSLM has a lower genPPL for spoken continuations than speech-only baselines, approaching the performance of a text-leading model without requiring the prompt transcript at inference. These results suggest that text provides a more effective target for learning the linguistic content of future speech, since the continuation is learned directly in text rather than through the speech prediction objective.

The explicit text prediction also provides a direct interface for distillation from the pretrained text LM. We distill the frozen text backbone into TGSLM by minimizing the KL divergence between their next-token distributions at text-prediction positions and refer to the resulting model as TGSLM-distill (combined with text replay, Sec.~\ref{sec:replay}). Distillation improves factual performance under both text-to-text and speech-to-speech evaluation, showing that capabilities from the pretrained text model can be better maintained during speech adaptation.


Our main contributions are:
\begin{itemize}
\item We study a speech-to-speech pretraining formulation (TGSLM) in which we predict future text before its corresponding speech, while the input remains speech-only at inference. Under matched training conditions, this formulation substantially improves spoken continuation modeling over speech-only and interleaved baselines and approaches an inner-monologue model with an ASR-transcribed prompt text. 

\item We distill a pretrained text LM into TGSLM's intermediate text predictions. Distillation improves factual performance in both speech-to-speech and text-to-text evaluation, indicating better retention of the pretrained LM's capabilities during speech adaptation. To the best of our knowledge, this is the first application of distillation from text LMs to speech-to-speech LMs. 

\item Additionally, we evaluate sensitivity to paralinguistic cues and find that models with internal text prediction and distillation remain comparably sensitive to emotional prosody when prompted with speech sharing the same linguistic content.
\end{itemize}

\section{Related work}
\subsection{Pure Speech Language Models}

Pure SLMs learn to model speech without using text as part of the objective. Early work on generative spoken language modeling showed that speech can be modeled autoregressively by discretizing self-supervised speech representations into ``semantic" tokens \citep{lakhotia2021generative}.\footnote{This term refers to tokens that represent the linguistic (not only semantic) content.} TWIST initializes a pure SLM from a pretrained text LM, which improves performance on linguistic evaluations\citep{hassid2023textually}. SLMs over continuous speech features combine autoregressive generation with a generative head, using conditional flow matching \citep{chou2025flow} or diffusion \citep{rouard2026continuous}, to jointly learn linguistic and acoustic information. Llama-Mimi flattens semantic tokens and acoustic tokens (the remaining RVQ codebooks, which capture acoustic detail needed for waveform reconstruction) into a single autoregressive sequence \citep{sugiura2025llama}. 
These approaches provide ways of learning pure speech models, either without text supervision or with text pretraining, but their performance on tasks involving linguistic content remains limited~\citep{chou2026quantifying}.  This motivates introducing {\it explicit} text supervision during speech pretraining.

%

\subsection{Speech-Text Language Models}

Speech-text language models directly predict both speech and text sequences.
Interleaved SLMs alternate between speech and text spans of the same utterance, typically switching modalities at word boundaries \citep{chou2023toward,nguyen2025spirit}. In this formulation, text provides an additional training signal, but is not used explicitly when generating a speech continuation.

Another family is inner-monologue modeling, where text prediction precedes and guides speech generation. Spectron follows a speech prompt $\rightarrow$ prompt transcription $\rightarrow$ text continuation $\rightarrow$ speech continuation structure \citep{nachmani2024spoken}. Moshi uses a word-aligned inner monologue in which text precedes the corresponding generated audio \citep{defossez2024moshi}. FLM-Audio uses sentence-level monologues that let text run ahead of speech~\citep{yao2025flm}. Thinker-Talker architectures such as Qwen2.5-Omni~\citep{Qwen2.5-Omni} modularize a related text-before-speech dependency, with a separate Talker generating speech from the Thinker's text and hidden representations.

TGSLM combines complementary aspects of interleaved and inner-monologue models.
At word boundaries, rather than switching between speech and text spans, TGSLM
transitions to text-guided generation: similar to inner-monologue models, it first
predicts the text and then generates the corresponding speech realization.

Moshi~\citep{defossez2024moshi} and FLM-Audio~\citep{yao2025flm} use related speech-input, text-guided speech generation structures. However, they use it in dialogue post-training, where it is learned together with conversational behaviors such as turn-taking and full-duplex interaction. We instead use it in speech pretraining,  isolating its effect on linguistic modeling from dialogue training.

\subsection{Knowledge Transfer from Text Language Models}

As text language models acquire broad linguistic and factual capabilities, a natural question is how to transfer these capabilities to speech models.
Most prior distillation methods are speech-input, text-output models. BLSP-KD, DiVA, and SALAD follow this general paradigm, using speech-text pairs to align speech-conditioned predictions or representations with their text-conditioned counterparts~\citep{wang2024blsp,held2025distilling,cuervo2026closing}.
SALAD further studies forgetting of text capabilities during speech adaptation and combines cross-modal distillation with text replay.

Speech-to-speech models, however, produce no text distribution to distill into.
Align-SLM transfers semantic knowledge from a text LLM through preference optimization: it generates multiple speech continuations, scores their semantic quality using an LLM, and constructs preference pairs for DPO~\citep{lin2025align}.
This allows knowledge transfer without introducing text prediction, but the teacher signal is mediated through generated speech and sequence-level preferences.

TGSLM instead introduces an explicit text prediction interface within a speech-to-speech model.
This allows the pretrained text LM to directly supervise TGSLM's next-token predictions, while the model still takes speech as input and produces speech as its final output.
\section{Method}

\subsection{Speech representation and generation objective}

We follow the speech representation and generation objective of
Flow-SLM~\citep{chou2025flow,chou2026quantifying}.
Given an input waveform, the Mimi tokenizer~\citep{defossez2024moshi}
produces synchronized discrete and continuous speech representations.
We use the first Mimi RVQ~\citep{zeghidour2021soundstream} codebook as discrete semantic tokens
$\mathbf{z}=(z_1,\ldots,z_M)$ and the pre-quantized Mimi features as
continuous acoustic representations
$\mathbf{x}=(x_1,\ldots,x_M)$.

At each speech position $m$, the Transformer~\citep{vaswani2017attention} encodes the preceding context
into a hidden state $h_m$.
A discrete speech head autoregressively predicts the next $k=4$ semantic tokens,
$p_\theta(z_{m:m+k-1}\mid h_m)
=
\prod_{i=0}^{k-1}
p_\theta^{(i)}
\left(
z_{m+i}\mid h_m,z_{m:m+i-1}
\right),$
with corresponding negative log-probability loss $\mathcal{L}_{\mathrm{discrete}}$.

To model acoustic information beyond the discrete semantic tokens, a conditional
flow-matching (CFM)~\citep{lipman2022flow} head models
$p_\theta(x_m\mid h_m,z_{m:m+k-1})$,
conditioning the continuous representation on both the Transformer state and
the predicted semantic tokens.
Following~\citep{chou2026quantifying}, we use the $x$-prediction
parameterization~\citep{li2026back} and optimize the corresponding CFM loss
$\mathcal{L}_{\mathrm{CFM}}$.
The full speech-generation objective is
\begin{equation}
\mathcal{L}_{\mathrm{speech}}
=
\mathcal{L}_{\mathrm{discrete}}
+
\mathcal{L}_{\mathrm{CFM}}.
\end{equation}
\subsection{Text-Guided Sequence Construction}

Fig.~\ref{fig:arch} illustrates the sequence construction.
We first obtain word-level speech-text alignments using a Whisper-based alignment method~\citep{yeh2025whisper}.
For each training utterance, we randomly sample a word boundary and insert a \texttt{<switch>} token at that position.
Speech preceding the switch is used only as context and incurs no loss.

After the switch, generation proceeds word by word.
Let \(T_i\) denote the sequence of text tokens corresponding to the \(i^\textrm{th}\) word after the switch point, and let \(S_i\) denote its speech segment.
The sequence is constructed as
\begin{equation}
S_{\mathrm{context}}
\rightarrow \texttt{<switch>}
\rightarrow T_1
\rightarrow S_1
\rightarrow T_2
\rightarrow S_2
\rightarrow \cdots .
\end{equation}
Thus, each continuation word is first predicted as text tokens and then generated as its corresponding speech segment before the model proceeds to the next word.

At text positions, we apply the standard autoregressive cross-entropy loss to the tokens in \(T_i\), conditioned on the full preceding speech-text context.
At speech positions, we apply \(\mathcal{L}_{\mathrm{speech}}\).
Only the output head corresponding to the current modality is trained at each position.

To mark the end-of-word boundaries between text and speech generation, we introduce
\texttt{<t-eow>} and \texttt{<s-eow>} tokens.
The model predicts \texttt{<t-eow>} after the final text token of each word and
\texttt{<s-eow>} after the corresponding speech segment.
We treat \texttt{<s-eow>} as an additional symbol in the vocabulary of the first
discrete-token prediction \(p_\theta^{(0)}\).


\subsection{Knowledge Distillation from the Text LM}\label{sec:replay}

The explicit text predictions provide a direct interface for knowledge distillation from the pretrained text LM.
We use the original backbone LM, before speech adaptation, as a frozen teacher.
At each text-prediction position, the frozen teacher conditions on the ground-truth text corresponding to the speech context up to that position, while the student is conditioned on its speech context (prompt) and the speech-text history after the switch token.
Our distillation loss is the forward KL divergence between their next-token distributions~\citep{hinton2015distilling}:
\begin{equation}
\mathcal{L}_{\mathrm{KD}}
=
\mathbb{E}
\left[\frac{1}{T}\sum_{t=1}^T
D_{\mathrm{KL}}
\left(
p_{\mathrm{teacher}}(y_t \mid c^{\mathrm{text}}_{<t})
\;\Vert\;
p_{\theta}(y_t  \mid c^{\mathrm{st}}_{<t})
\right)
\right],
\end{equation}
where $c^{\mathrm{text}}_{<t}$ and $c^{\mathrm{st}}_{<t}$ denote the teacher and student contexts before timestep $t$, and $y_t$ denotes the text token at step $t$. $T$ denotes the text token sequence length. 
We interpolate the standard text prediction loss and the distillation loss. The full training objective is

\begin{equation}
\mathcal{L}
=
(1-\lambda)\mathcal{L}_{\mathrm{text}}
+
\lambda\mathcal{L}_{\mathrm{KD}}
+
\mathcal{L}_{\mathrm{speech}},
\end{equation}
where $\lambda$ controls the distillation strength. Setting $\lambda=0$ recovers training without distillation.
We also train with text-only batches to mitigate forgetting. For replay-enabled models, a text-only batch is sampled with probability
0.2 and optimized using the standard next-token prediction objective.

\subsection{Inference}\label{sec:inference}

At inference time, a prompt is provided as speech context.
We append \texttt{<switch>} immediately after the speech prompt to trigger text-guided generation.
The model then alternates between text and speech generation, constructing the word-level text-speech alignment online.

Word boundaries are determined by a greedy criterion: A segment ends when its end-of-word symbol becomes the highest-ranked prediction. For speech, the current segment ends when
\begin{equation}\label{eq:align}
\arg\max_{v} p_\theta^{(0)}(v\mid h_m) = \texttt{<s-eow>},
\end{equation}
and the text segment ends when \texttt{<t-eow>} is the highest-ranked text token.
The model generates text tokens until \texttt{<t-eow>}, switches to speech generation, where it samples discrete speech tokens and generates the corresponding continuous representations until \texttt{<s-eow>}, and then returns to text generation for the next word. This process continues until the text stream produces an \texttt{<EOS>} token, after which generation terminates upon completing the corresponding final speech segment.

For likelihood-based evaluation (StoryCloze), we score only text-token positions in the continuations. Boundary handling is described in Appendix~\ref{sec:likelihood}.

\section{Experiments}

\subsection{Data and Training}\label{sec:data}

We train on the English subset of Emilia~\citep{he2024emilia}, adding the English subset of Emilia-YODAS for the scaled (TGSLM-Large) model. We filter alignments for quality (Appendix~\ref{sec:quality}). Training and decoding hyperparameters are in Appendix~\ref{sec:training_hp_v2}. We use FineWeb 10B subset for text replay~\citep{penedo2024fineweb}.

For evaluation, we synthesize speech for StoryCloze, Known-1000, and a dataset we create called Gemini-Emo to test emotion sensitivity.
For StoryCloze, following prior work~\citep{lu2026latent,cuervo2026closing,maimon2025scaling}, we synthesize the story context and candidate continuations with Kokoro-TTS using the \texttt{af\_heart} voice.\footnote{\url{https://huggingface.co/hexgrad/Kokoro-82M}}
For Known-1000, we synthesize the factual prefixes using NVIDIA Magpie TTS with the \texttt{Sofia} voice\footnote{\url{https://huggingface.co/nvidia/magpie_tts_multilingual_357m}}.

To test emotion sensitivity, we construct Gemini-Emo: 96 triplets of happy, sad, and angry utterances synthesized with emotional prosody, sharing a linguistically neutral spoken prefix that serves as the prompt (details and an example in Appendix~\ref{sec:gemini_prompt}).

To avoid tokenizer artifacts at the end of prompts, we trim trailing silence and pad each waveform to a multiple of the tokenizer frame duration (details in Appendix~\ref{sec:vad}).

\subsection{Model Architecture}

All models are initialized from pretrained Qwen3 base LMs~\citep{yang2025qwen3}: Qwen3-0.6B for TGSLM variants and Qwen3-1.7B for the scaled TGSLM-Large. Unless otherwise specified, \textbf{TGSLM} is trained without distillation or text replay, and \textbf{TGSLM-distill} uses both. We report $\lambda$ when it varies.

\subsection{Evaluation}\label{sec:eval}

\paragraph{Generative perplexity (genPPL).}
We use genPPL to evaluate the linguistic quality of open-ended speech continuations, following~\citep{chou2026quantifying,maimon2025slamming}.
Given a speech prompt, each model generates five continuations.
We transcribe the generated speech with Whisper \texttt{small.en}~\citep{radford2023robust} and score the continuation transcript with OLMo-3 7B~\citep{team2025olmo}, conditioned on the prompt transcript.
Lower genPPL indicates that the generated continuation is more likely under the reference text LM. We score only the continuation tokens so that genPPL reflects continuation quality rather than the likelihood of the given prompt.
To identify the continuation in the ASR transcription, we locate the prompt boundary by finding the minimum character edit distance between the ASR transcription and the prompt text, following~\citep{chou2026quantifying}. We use the intersection of utterances with non-empty continuations across all compared models, so that every model is evaluated on the same subset.

\paragraph{Auto-BLEU.}
Following prior work~\citep{lakhotia2021generative,lin2025align,maimon2025slamming}, we additionally report bigram auto-BLEU to measure repetition. Auto-BLEU computes the fraction of generated N-grams ($N=2$) that appear more than once within the utterance. This complements generative perplexity, since repetition can result in an artificially low perplexity under a reference language model~\citep{holtzman2019curious,welleck2019neural,zhang2026low}.

\paragraph{StoryCloze.}
We evaluate spoken StoryCloze~\citep{maimon2025slamming,lu2026latent,hassid2023textually} by comparing candidate likelihoods, computed over text-token predictions only (Sec.~\ref{sec:inference}).

\paragraph{Factual knowledge.}
GenPPL averages over all continuation tokens and may not fully reflect whether specific factual knowledge is retained.
We evaluate factual knowledge using Known-1000~\citep{meng2022locating},\footnote{\url{https://rome.baulab.info/data/dsets/known_1000.json}} which tests factual completion through open-ended continuation and more closely matches our generation setting than question-answering benchmarks.
For each prompt, we generate five spoken continuations and transcribe them with Whisper \texttt{small.en}.
Following Spectron~\citep{nachmani2024spoken}, a sample is considered correct if the answer appears in its transcription.
We count an example as correct under Pass@\(K\) if at least one of the first \(K\) generated samples is correct.

\paragraph{Emotion sensitivity.}
We evaluate emotion sensitivity on Gemini-Emo using the prompt triplets described in Sec.~\ref{sec:data}.
For each generated sample, we apply the emotion2vec-plus-large~\citep{ma2024emotion2vec} classifier to the generated continuation, with happy, angry, and sad emotion classes.


\paragraph{Acoustic quality and speaker similarity.}
We evaluate acoustic quality using UTMOS-v2~\citep{baba2024utmosv2}.
We measure speaker similarity between the speech prompt and generated continuation using WavLM-large-TDNN, following~\citep{he2024emilia,le2023voicebox}.\footnote{\url{https://github.com/microsoft/UniSpeech/tree/main/downstreams/speaker_verification}}

\section{Results}
\subsection{Text-guided speech generation improves genPPL}


\begin{table}[t]
\centering
\small
\setlength{\tabcolsep}{2.2pt}
\renewcommand{\arraystretch}{1.06}

\caption{
Generation-based metrics on the Emilia test set: genPPL, bigram auto-BLEU (BLEU), speaker similarity (Spk), and UTMOS-v2 (UTMOS)~\citep{baba2024utmosv2}; see Sec.~\ref{sec:eval} for details. For the \textbf{Whisper prompt transcript} row in text-leading references, the model is conditioned on a transcript of the speech prompt produced by Whisper \texttt{small.en}. \textbf{TGSLM-Large-distill} uses $\lambda=0.5$ with text replay. All rows are evaluated on the same subset: a shared subset of 1,325 prompts (of 1,500) with non-empty continuations from every model, including external models in Appendix~\ref{sec:full}, covering 4,957 of 7,500 continuations.
}
\label{tab:genppl}

\begin{tabularx}{\linewidth}{
@{}
>{\raggedright\arraybackslash}X
>{\centering\arraybackslash}p{0.145\linewidth}
>{\centering\arraybackslash}p{0.065\linewidth}
>{\centering\arraybackslash}p{0.085\linewidth}
r
r
r
r
@{}
}
\toprule
\textbf{Model}
& \textbf{Direction}
& \textbf{Params}
& \textbf{Data}
& \textbf{GenPPL $\downarrow$}
& \textbf{BLEU$\downarrow$}
& \textbf{Spk  $\uparrow$}
& \textbf{UTMOS $\uparrow$} \\
\midrule

\rowcolor{gray!10}
\textbf{Ground truth}
    & --
    & --
    & --
    & 43.8
    & 6.0
    & 0.74
    & 2.82 \\

\addlinespace[2pt]

\rowcolor{gray!15}
\multicolumn{8}{@{}l@{}}{
    \textbf{Speech-only baselines (no text guidance)}
} \\

Speech-only
    & S $\rightarrow$ S
    & 0.8B
    & 21.3k h
    & 151.7
    & 4.3
    & 0.42
    & 2.56 \\

Interleaved
    & S $\rightarrow$ S
    & 0.8B
    & 21.3k h
    & 164.1
    & 5.4
    & 0.37
    & 2.36 \\

\addlinespace[2pt]

\rowcolor{blue!18}
\makebox[0pt][l]{\textbf{TGSLM variants}}
    & & & & & & & \\

TGSLM
    & S $\rightarrow$ (T, S)
    & 0.8B
    & 21.3k h
    & 58.9
    & 3.9
    & 0.36
    & 2.41 \\

\quad + self-generated prompt text
    & (S, T) $\rightarrow$ (T, S)
    & 0.8B
    & 21.3k h
    & 62.2
    & 3.8
    & 0.38
    & 2.40 \\

\quad + distillation ($\lambda=0.25$)
    & S $\rightarrow$ (T, S)
    & 0.8B
    & 21.3k h
    & 56.9
    & 4.3
    & 0.36
    & 2.41 \\

\quad + distillation + text replay
    & S $\rightarrow$ (T, S)
    & 0.8B
    & 21.3k h
    & 51.0
    & 4.9
    & 0.38
    & 2.44 \\

TGSLM-Large-distill
    & S $\rightarrow$ (T, S)
    & 1.9B
    & 135k h
    & 45.0
    & 5.7
    & 0.45
    & 2.68 \\

\addlinespace[2pt]

\rowcolor{orange!18}
\makebox[0pt][l]{\textbf{Text-leading references}}
    & & & & & & & \\

Oracle prompt text
    & (T, S) $\rightarrow$ (T, S)
    & 0.8B
    & 21.3k h
    & 40.7
    & 5.4
    & 0.42
    & 2.50 \\

Whisper prompt transcript
    & (T, S) $\rightarrow$ (T, S)
    & 0.8B
    & 21.3k h
    & 57.1
    & 6.5
    & 0.41
    & 2.50 \\

\addlinespace[2pt]

\bottomrule
\end{tabularx}

\vspace{1mm}
\end{table}

As shown in Table~\ref{tab:genppl}, TGSLM reduces genPPL from 151.7 for the speech-only baseline to 58.9 at the same model and data scale, while
interleaved training does not improve over the speech-only baseline.\footnote{Under matched update counts, the interleaved model's sequence budget is shared between speech and text tokens, so it sees roughly half the speech per update compared with the speech-only and TGSLM models.} 
Distillation and text replay further reduce genPPL to 56.9 and 51.0,
respectively, indicating that transferring the text LM's distribution also improves open-ended continuation. Auto-BLEU remains relatively stable for TGSLM variants and speech-only baselines, indicating that the genPPL improvements do not come from more repetitive continuations. 

We also test whether explicitly predicting the speech prompt text provides further benefit.
In the \textbf{+ self-generated prompt text} row in Table \ref{tab:genppl}, the prompt is represented as an alternating sequence $S_1 \rightarrow T_1 \rightarrow S_2 \rightarrow T_2 \rightarrow \cdots$,
where each speech word segment is followed by its generated text transcription.
After the model consumes the prompt, generation proceeds with standard TGSLM text-guided continuation (text before speech).
At inference, where the prompt transcript is unavailable, the model consumes the speech prompt sequentially and identifies word boundaries online using the same \texttt{<s-eow>} criterion as in Eq. \ref{eq:align}.
This variant does not improve over TGSLM, suggesting that explicitly recovering the prompt text is not necessary to obtain the benefits of text-guided continuation.

We also compare against two text-leading references, conditioned on either the ground-truth or the Whisper-transcribed prompt text. With oracle prompt text, the text-leading model achieves lower genPPL (40.7) than TGSLM (58.9). However, replacing the oracle transcript with the Whisper-generated transcript degrades performance. TGSLM approaches the text-leading model conditioned on a Whisper transcript (58.9 vs.~57.1) without requiring prompt transcription.
This suggests that explicitly providing prompt text is not necessary to obtain most of the benefit of text-guided generation. The more important factor is predicting future text before generating its corresponding speech.

For acoustic metrics, the controlled TGSLM models show slightly lower speaker similarity and UTMOS-v2 than the speech-only baseline, consistent with the tradeoff observed in prior work~\citep{chou2026quantifying}.
The differences are modest, and the scaled TGSLM-Large-distill model recovers both metrics.


\subsection{Distillation improves factual knowledge retention}
\begin{figure}[h]
\includegraphics[width=\linewidth]{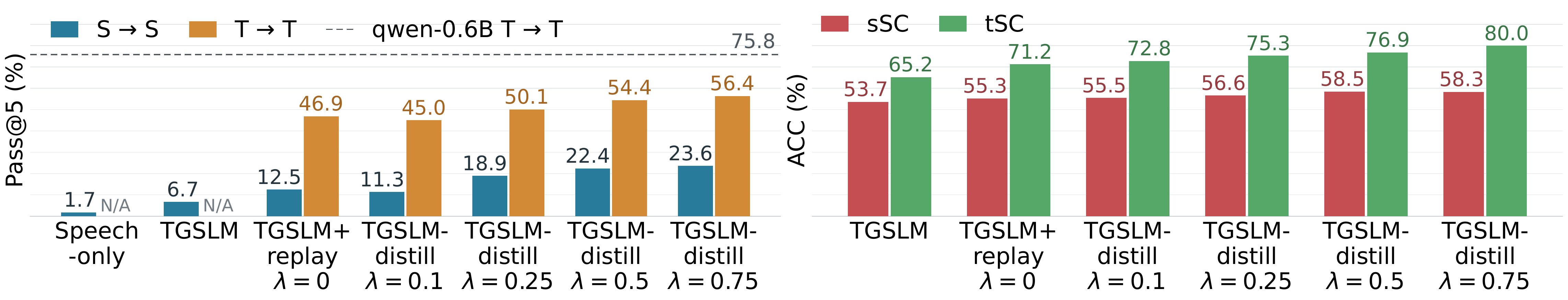}

    \caption{Ablation of text replay and distillation weight $\lambda$. TGSLM uses neither. TGSLM + replay adds text replay only ($\lambda=0$); TGSLM-distill adds both. Left: Pass@5 (\%) on Known-1000. Right: accuracy (\%) on spoken (sSC) and text (tSC) StoryCloze~\citep{hassid2023textually}.}
    \label{fig:distill_hp}
\end{figure}


\newcommand{\ptgray}[1]{\textcolor{black!50
}{#1}}

\begin{table*}[t]
\centering
\small
\setlength{\tabcolsep}{4pt}
\renewcommand{\arraystretch}{1.06}

\caption{
Known-1000 evaluation.
Pass@\(K\)(\%) measures whether the answer appears in at least one of \(K\) sampled continuations.
Direction S $\rightarrow$ (T, S) denotes a speech prompt followed by a text-guided speech continuation.
\textbf{TGSLM-Large-distill} uses a distillation weight of $\lambda=0.5$ and text replay.
}
\label{tab:known1000}

\begin{tabularx}{\textwidth}{
@{}
>{\raggedright\arraybackslash}X
>{\centering\arraybackslash}p{0.17\textwidth}
>{\centering\arraybackslash}p{0.08\textwidth}
>{\centering\arraybackslash}p{0.10\textwidth}
r
r
@{}
}
\toprule
\textbf{Model}
& \textbf{Direction}
& \textbf{Params}
& \textbf{Data}
& \textbf{Pass@1}
& \textbf{Pass@5} \\
\midrule

\rowcolor{gray!15}
\multicolumn{6}{@{}l@{}}{\textbf{Text-LM reference}} \\
Qwen3-0.6B backbone~\citep{yang2025qwen3}
    & T $\rightarrow$ T
    & 0.6B
    & --
    & 46.8
    & 77.9 \\
Qwen3-1.7B backbone~\citep{yang2025qwen3}
    & T $\rightarrow$ T
    & 1.7B
    & --
    & 60.8
    & 86.6 \\

\addlinespace[2pt]

 \rowcolor{blue!18}
 \multicolumn{6}{@{}l@{}}{\textbf{Our models}} \\
TGSLM-distill
    & S $\rightarrow$ (T, S)
    & 0.8B
    & 45k h
    & 9.1
    & 22.4 \\
TGSLM-Large-distill
    & S $\rightarrow$ (T, S)
    & 1.9B
    & 135k h
    & 16.8
    & 37.9 \\


\addlinespace[2pt]

\rowcolor{gray!15}
\multicolumn{6}{@{}l@{}}{\textbf{External pretrained models}} \\

\rowcolor{orange!10}
\multicolumn{6}{@{}l@{}}{
    \textbf{\textcolor{orange!35!black}{With text supervision}}
} \\

SpiritLM Base~\citep{nguyen2025spirit}
    & S $\rightarrow$ S
    & 7B
    & 570k h
    & 8.6
    & 23.5 \\

    & S $\rightarrow$ T
    &
    &
    & 15.1
    & 31.8 \\


MiMo-Audio-7B-Base~\citep{xiaomi2025mimo}
    & S $\rightarrow$ (T, S)
    & 7B
    & \mbox{$>$100M h}
    & 26.0
    & 49.5 \\

\addlinespace[1pt]

\rowcolor{green!10}
\multicolumn{6}{@{}l@{}}{
    \textbf{\textcolor{green!45!black}{Without text supervision}}
} \\

SLAM-Large~\citep{maimon2025slamming}
    & S $\rightarrow$ S
    & 1.3B
    & 81.0k h
    & 1.2
    & 5.5 \\

Llama-Mimi-1.3B~\citep{sugiura2025llama}
    & S $\rightarrow$ S
    & 1.3B
    & 240k h
    & 0.2
    & 0.6 \\

Llama-Mimi-8B~\citep{sugiura2025llama}
    & S $\rightarrow$ S
    & 8B
    & 240k h
    & 0.9
    & 2.9 \\

\addlinespace[2pt]





\bottomrule
\end{tabularx}

\vspace{1mm}


\end{table*}
Fig. \ref{fig:distill_hp} shows the results of ablation over text replay and distillation weight.
On Known-1000 (left), text replay alone already improves factual performance relative to TGSLM, showing that maintaining exposure to text during speech adaptation mitigates part of the loss of pretrained knowledge. Distillation provides an additional gain beyond replay, with performance
generally improving as the distillation weight grows. A similar trend is observed on StoryCloze (right).


Importantly, these improvements are not limited to speech-to-speech generation.
Distillation also improves text-to-text performance, even though the distillation loss during speech training is applied to text predictions conditioned on speech context.
This suggests that distillation is not merely teaching the model to map speech contexts to better next-token predictions. It also constrains
speech adaptation in a way that better preserves the capabilities of the original text backbone.
A similar effect is observed in SALAD~\citep{cuervo2026closing}, where cross-modal distillation reduces the degradation of the original text capabilities after speech adaptation.

Table \ref{tab:known1000} compares the scaled TGSLM with distillation and text replay against external models. Scaling up the distilled model improves Known-1000 Pass@5 from 22.4\% to 37.9\%.
Despite using a smaller language model and less speech data, TGSLM-Large-distill exceeds SpiritLM Base on both speech-to-speech and speech-to-text results.\footnote{This comparison is not controlled for data distribution, architecture, base LM, or tokenizers.} Together with the controlled ablation in Fig.~\ref{fig:distill_hp}, this suggests that distillation is an efficient way to retain factual knowledge in speech-to-speech models.

The contrast is larger for pretrained models without text supervision.
SLAM-Large and the Llama-Mimi models obtain Pass@5 below 6\%, and increasing the size of Llama-Mimi from 1.3B to 8B parameters yields only a small gain (0.6\% to 2.9\%). This suggests that scaling pretraining without text supervision is a less efficient way to retain factual knowledge than text-guided generation with distillation. 



\subsection{Distillation has limited impact on emotion sensitivity}

\begin{table*}[t]
\centering
\small
\setlength{\tabcolsep}{5pt}
\renewcommand{\arraystretch}{1.08}

\newcommand{\gemoCell}[2]{ $#1\,{\color{gray!60}\pm #2}$ }
\newcommand{\gemoRef}[2]{ $#1\,{\color{gray!60}\pm #2}$ }

\caption{
Gemini-Emo three-way emotion classification accuracy (\%, higher is better) on continuation using
emotion2vec-plus-large~\citep{ma2024emotion2vec}.
We report 95\% percentile bootstrap confidence intervals from 1000 resamples of the 96 triplets. $\pm h$ denotes the half-width interval.
Generation failures and empty continuations are counted as incorrect.
}
\label{tab:gemini_emo}

\begin{tabularx}{\textwidth}{
@{}
>{\raggedright\arraybackslash}X
c
c
c
c
@{}
}
\toprule
\textbf{Model}
& \textbf{Happy}
& \textbf{Sad}
& \textbf{Angry}
& \textbf{Macro} \\
\midrule

\rowcolor{gray!12}
\textbf{Original prefix only}
    & \gemoRef{64.6}{9.4}
    & \gemoRef{92.7}{4.7}
    & \gemoRef{77.1}{8.3}
    & \gemoRef{78.1}{4.3} \\

\addlinespace[2pt]

\rowcolor{blue!18}
\multicolumn{5}{@{}l}{\textbf{Our models}} \\

Speech-only
    & \gemoCell{54.2}{5.2}
    & \gemoCell{76.9}{4.3}
    & \gemoCell{36.7}{4.5}
    & \gemoCell{55.9}{2.7} \\

TGSLM ($\lambda=0$)
    & \gemoCell{57.7}{4.9}
    & \gemoCell{72.1}{4.6}
    & \gemoCell{43.3}{5.0}
    & \gemoCell{57.7}{2.8} \\

TGSLM-distill ($\lambda=0.5$)
    & \gemoCell{60.0}{4.4}
    & \gemoCell{77.3}{4.3}
    & \gemoCell{34.8}{4.7}
    & \gemoCell{57.4}{2.6} \\

TGSLM-Large-distill ($\lambda=0.5$)
    & \gemoCell{56.0}{4.6}
    & \gemoCell{75.4}{4.3}
    & \gemoCell{43.5}{5.4}
    & \gemoCell{58.3}{2.6} \\

\addlinespace[2pt]


\rowcolor{orange!12}
\multicolumn{5}{@{}l}{
    \textbf{\textcolor{orange!55!black}{External models with limited acoustic information}}
} \\

SLAM-Large~\citep{maimon2025slamming}
    & \gemoCell{65.4}{4.2}
    & \gemoCell{34.4}{4.6}
    & \gemoCell{1.9}{1.1}
    & \gemoCell{33.9}{1.9} \\

SpiritLM Base~\citep{nguyen2025spirit}
    & \gemoCell{75.4}{4.4}
    & \gemoCell{23.8}{4.3}
    & \gemoCell{2.7}{1.6}
    & \gemoCell{34.0}{1.8} \\

SpiritLM Expressive~\citep{nguyen2025spirit}
    & \gemoCell{66.0}{4.9}
    & \gemoCell{76.7}{4.0}
    & \gemoCell{9.2}{2.9}
    & \gemoCell{50.6}{2.2} \\

\rowcolor{green!12}
\multicolumn{5}{@{}l}{
    \textbf{\textcolor{green!40!black}{External models with richer acoustic information}}
} \\

MiMo-Audio Base~\citep{xiaomi2025mimo}
    & \gemoCell{73.1}{4.2}
    & \gemoCell{45.8}{5.2}
    & \gemoCell{25.6}{4.6}
    & \gemoCell{48.2}{2.5} \\

Llama-Mimi 1.3B~\citep{sugiura2025llama}
    & \gemoCell{46.5}{4.9}
    & \gemoCell{87.7}{4.0}
    & \gemoCell{45.0}{4.8}
    & \gemoCell{59.7}{2.3} \\

Llama-Mimi 8B~\citep{sugiura2025llama}
    & \gemoCell{46.0}{4.8}
    & \gemoCell{88.5}{3.2}
    & \gemoCell{46.5}{5.0}
    & \gemoCell{60.3}{2.3} \\

\addlinespace[2pt]


\bottomrule
\end{tabularx}

\vspace{1mm}
\end{table*}

A natural question is whether distillation reduces sensitivity to paralinguistic information as the model is trained to mimic the behavior of a text LM.
We evaluate this using Gemini-Emo together with an emotion classifier, with results shown in Table \ref{tab:gemini_emo}.
TGSLM and TGSLM-distill achieve similar emotion classification accuracy, showing limited degradation in emotion sensitivity from distillation in this evaluation.
TGSLM-Large-distill performs slightly worse than Llama-Mimi, but remains broadly comparable.

Models based on more limited acoustic representations
(\textbf{External models with limited acoustic information} row)
show substantially more class-imbalanced emotion predictions,
which we attribute to the loss of acoustic information during tokenization, as these models use either only semantic tokens or semantic tokens plus pitch tokens.

We view this as a preliminary study of paralinguistic sensitivity rather
than a comprehensive evaluation of the generated continuation, as Gemini-Emo
uses a single emotion classifier and primarily probes whether acoustic emotion
cues in the prompt are reflected in the generated speech. It does not test whether distillation toward a text-only teacher shifts the lexical content toward neutral content.



\section{Conclusion}

We introduce TGSLM, a speech-to-speech LM that uses internally predicted text to guide speech generation while maintaining speech-only input and output at inference.
The TGSLM internal text construction substantially improves generative perplexity and provides an interface for text-LM distillation, which improves factual knowledge retention in both speech-to-speech and text-to-text evaluation. An emotion sensitivity evaluation suggests that distillation has limited impact on this aspect of generation.

TGSLM currently relies on word-level speech-text alignments to construct
the training sequences.
We have not studied whether this alignment requirement can be eliminated.
One possible direction is to predict text and speech at the sentence level,
which could provide text guidance without fine-grained alignment.
We also focus on pretraining in this work and leave post-training, such as
instruction tuning or dialogue adaptation, for future work.

\section*{AI use statement}

In this work, we used  AI tools to generate synthetic evaluation data, assist with implementation, and with experimental code. In particular, we used AI to construct synthetic text and speech data as described in the paper, and coding agents to assist with implementation and debugging.
We also used AI tools to help draft and edit the paper for clarity and readability, as well as for brainstorming and code-related assistance.
The authors reviewed and tested AI-assisted code, and verified all experimental results. The authors reviewed and revised the AI-assisted manuscript text. We take responsibility for the final content of this work.

\bibliography{iclr2027_conference_conf}
\bibliographystyle{iclr2027_conference}
\clearpage
\appendix
\section{Appendix}

\subsection{Alignment filtering rules}\label{sec:quality}
First, we require the final aligned word to reach at least 80\% of the audio duration. Second, we require that no more than 5\% of aligned words have effectively zero duration. We exclude utterances whose transcripts exceed 7.5 Qwen3-0.6B text tokens per second.

\subsection{likelihood measure}\label{sec:likelihood}
For likelihood-based evaluation, we consume the given text and speech continuation using the same alternating procedure and accumulate likelihood only at text-token positions.
Because the text-speech alignment is reconstructed online, the strict top-1 boundary criterion can occasionally fail to trigger enough speech boundaries, leaving some text tokens unscored.
Only in this case, we use a fallback boundary criterion and accept \texttt{<s-eow>} whenever it appears among the top-$3$ predictions of the first discrete speech-token head.
This fallback is used only for likelihood-based scoring (StoryCloze) and is not applied during generation.

\subsection{VAD protocol for known-1000}\label{sec:vad}
We observe that abruptly ending a synthesized waveform can introduce tokenizer artifacts in the final speech frames, creating an artificial discontinuity at the end of the input.
To reduce this effect, we first trim trailing silence using voice activity detection (VAD)~\citep{hwang2023torchaudio} and then pad each waveform so that its duration is a multiple of the tokenizer frame duration.
We use a frame duration of 80\,ms for Mimi and 40\,ms for HuBERT-based~\citep{hsu2021hubert} representations~\citep{maimon2025slamming,nguyen2025spirit}.
\subsection{Gemini-Emo Prompts}\label{sec:gemini_prompt}
For Gemini-Emo, we first use Gemini-3.7-Flash to generate triplets of sentences expressing happy, sad, and angry emotions while sharing the same linguistically neutral prefix. Table \ref{tab:gemini-emo-example} shows an example.
We construct the shared prefix so that its text alone does not indicate the target emotion.
We then use Gemini-3.1-Flash-TTS-Preview to synthesize each sentence with the corresponding emotional prosody.
Detailed prompts are provided in Table \ref{tab:gemini-emo-prompt-templates_v2}.
We obtain word-level speech-text alignments using the same Whisper-based alignment method ~\citep{yeh2025whisper} and extract the prefix as the speech prompt for continuation generation.
We classify the full synthesized utterances with emotion2vec-plus-large~\citep{ma2024emotion2vec}, restricting the prediction to the happy, sad, and angry classes, and retain only triplets in which all three utterances are classified as their intended emotions.
This results in 96 triplets.
\begin{table}[h]
\caption{An example from Gemini-Emo (\texttt{item\_01}). Each sentence combines the shared prefix with the corresponding suffix and is synthesized in the indicated emotion. Evaluation uses only the cropped spoken prefix, whose words are identical across emotions.}
\label{tab:gemini-emo-example}
\centering
\small
\begin{tabularx}{\linewidth}{@{}lX@{}}
\toprule
\textbf{Shared prefix} & I noticed that the bread in the kitchen \\
\midrule
\textbf{Emotion} & \textbf{suffix} \\
\midrule
Happy & has baked into the most wonderful golden loaf! \\
Sad & has turned stale and completely moldy already today. \\
Angry & was left uncovered on the counter again! \\
\bottomrule
\end{tabularx}

\end{table}

\begin{table*}[t]
\centering
\caption{Gemini-Emo language-model and TTS prompt specification. The shared TTS row applies to all three emotion prompts. Each prompt inserts one emotion-specific director instruction between the scene and transcript sections.}
\label{tab:gemini-emo-prompt-templates_v2}
\small
\setlength{\tabcolsep}{3pt}
\renewcommand{\arraystretch}{1.08}
\begin{tabularx}{\textwidth}{@{}>{\raggedright\arraybackslash}p{0.16\textwidth}>{\raggedright\arraybackslash}p{0.18\textwidth}X@{}}
\toprule
\textbf{Prompt purpose} & \textbf{Model} & \textbf{Prompt template} \\
\midrule
Text generation
  & \texttt{gemini-3.7}\par\texttt{-flash}
  & \parbox[t]{\linewidth}{Create exactly 25 English speech-continuation stimulus items. For every item: write one shared prefix of 7--12 words as a natural, affect-neutral, syntactically unfinished conversational turn; write emotion-specific 5--14-word suffixes that complete that prefix; form each carrier as \texttt{PREFIX + `` '' + SUFFIX}; use varied, ordinary ASR-friendly language; and return the supplied item schema. The generation ran in batches through 500 items.} \\
\midrule
Shared TTS interface
  & \parbox[t]{\linewidth}{\texttt{gemini-3.1}\par\texttt{-flash-}\par\texttt{tts-preview}\\voice: \texttt{Schedar}}
  & \parbox[t]{\linewidth}{Synthesize speech for the transcript below.\par\medskip\textbf{AUDIO PROFILE:} Use one consistent adult speaker identity and a standard American English accent.\par\medskip\textbf{SCENE:} This is a natural conversation with one familiar listener in a quiet room.\par\medskip\textbf{DIRECTOR'S NOTES:} \texttt{<EMOTION-SPECIFIC DIRECTOR'S NOTES>}\par\medskip Speak the transcript exactly once and verbatim. Do not read these directions. Do not add, remove, or replace words. Do not laugh, cry, sigh, gasp, whisper, shout, or add nonverbal sounds.\par\medskip\textbf{TRANSCRIPT:} \texttt{<TRANSCRIPT>}} \\
\midrule
Happy director's notes
  &
  & \parbox[t]{\linewidth}{Convey a small, private happiness, as if sharing good news with someone you trust. Keep an easy moderate pace and soft-to-medium energy. Let a subtle smile warm the tone and add light upward inflections. Sound understated but unmistakably pleased, not breathy, cute, or theatrical. Apply these cues from the first word, including the neutral lead-in. Keep the performance conversational, coherent, and moderately expressive rather than exaggerated.} \\
Sad director's notes
  &
  & \parbox[t]{\linewidth}{Convey strong genuine sorrow while trying to remain composed. Use a soft voice, slow measured pacing, lowered resonance, and pronounced downward intonation. Sound deeply sad but never cry, sob, or become theatrical. Apply these cues from the first word, including the neutral lead-in. Keep the performance conversational, coherent, and moderately expressive rather than exaggerated.} \\
Angry director's notes
  &
  & \parbox[t]{\linewidth}{Sound clearly angry but fully controlled. Use a firm tense tone, clipped articulation, strong stress, deliberate pacing, and moderately high energy. Address the listener directly without shouting. Apply these cues from the first word, including the neutral lead-in. Keep the performance conversational, coherent, and moderately expressive rather than exaggerated.} \\
\bottomrule
\end{tabularx}
\end{table*}

\subsection{training hyperparameters}\label{sec:training_hp_v2}
To reduce storage requirements, we pre-encode the Emilia-YODAS audio into Mimi RVQ tokens using 12 codebooks and decode them back to waveform on the fly during training.
Table \ref{tab:training-hyperparameters_v2} provides detailed training and decoding hyperparameters.

\begin{table}[h]
\caption{Training hyperparameters and decoding parameters. }
\label{tab:training-hyperparameters_v2}
\centering
\footnotesize
\begingroup
\setlength{\tabcolsep}{5pt}
\renewcommand{\arraystretch}{1.05}
\begin{tabularx}{\textwidth}{@{}p{0.27\textwidth}*{3}{>{\raggedright\arraybackslash}X}@{}}
\toprule
\textbf{Hyperparameter} & \textbf{Architecture ablations} & \textbf{Distillation ablations} & \textbf{TGSLM-Large-distill} \\
\midrule
\multicolumn{4}{@{}l}{\textit{Training}} \\
\midrule
Backbone & Qwen3-0.6B-Base & Qwen3-0.6B-Base & Qwen3-1.7B-Base \\
Optimizer updates & 40,000; 50,000 for model with text replay & 100,000 & 785,052 (665,000 + annealing) \\
Audio budget (s/GPU/update) & 480 & 540 & 240 \\
GPUs & 4 & 4 & 4 \\
Peak learning rate & $5\times10^{-4}$ & $5\times10^{-4}$ & $3\times10^{-4}$ \\
Warmup updates & 5,000 & 5,000 & 5,000 \\
End-of-cosine-decay / peak learning rate ratio & 0.1 & 0.1 & 0.1 \\
annealing & -- & -- & Linear annealing from $3\times10^{-5}$ to 0 \\
Bucket bin duration (s) & 2 & 2 & 2 \\ 
\midrule
FineWeb replay probability & 0; 0.2 for replay-enabled model & 0.2 & 0.2 \\
Replay tokens/GPU/update & 2,048  & 2,048 & 2,048 \\
\midrule
Optimizer & \multicolumn{3}{l}{AdamW8bit~\citep{dettmers20218}} \\
Adam $(\beta_1,\beta_2)$ & \multicolumn{3}{l}{$(0.9,0.98)$} \\
Adam $\epsilon$ & \multicolumn{3}{l}{$10^{-6}$} \\
Weight decay & \multicolumn{3}{l}{0.01} \\
Learning-rate schedule & \multicolumn{3}{l}{Linear warmup followed by cosine decay followed by optional linear annealing} \\
Training precision & \multicolumn{3}{l}{BF16 mixed precision} \\
\midrule
\multicolumn{4}{@{}l}{\textit{Shared decoding settings}} \\
\midrule
Sampling temperature & \multicolumn{3}{l}{0.8 for text, semantic tokens, and flow Gaussian} \\
Nucleus threshold $p$~\citep{holtzman2019curious} & \multicolumn{3}{l}{0.95} \\
Flow integration & \multicolumn{3}{l}{Euler, 32 steps} \\
CFG~\citep{ho2022classifier} scale & \multicolumn{3}{l}{0.3} \\
Mimi decoder quantizers & \multicolumn{3}{l}{16} \\
Samples per prompt & \multicolumn{3}{l}{5} \\
Maximum new speech & \multicolumn{3}{l}{10 s (Known-1000); 20 s (Emilia/Gemini-Emo)} \\
\bottomrule
\end{tabularx}
\endgroup

\vspace{2pt}
\begin{minipage}{\textwidth}

\end{minipage}
\end{table}

\subsection{full  results}\label{sec:full}
Full results for generation-based metrics for Emilia test set are in Table \ref{tab:genppl_full}, and known-1000 results with post-trained models are in Table \ref{tab:known1000_full}.


\begin{table}[t]
\centering
\small
\setlength{\tabcolsep}{2.2pt}
\renewcommand{\arraystretch}{1.06}

\caption{
Generation-based metrics on the Emilia test set: genPPL, bigram auto-BLEU (BLEU), speaker similarity (Spk), and UTMOS-v2 (UTMOS)~\citep{baba2024utmosv2}; see Sec.~\ref{sec:eval} for details. For the \textbf{Whisper prompt transcript} row in text-leading references, the model is conditioned on a transcript of the speech prompt produced by Whisper \texttt{small.en}. All rows are evaluated on the same subset: a shared subset of 1,325 prompts (of 1,500) with non-empty continuations from every model, covering 4,957 of 7,500 continuations. \textbf{TGSLM-Large-distill} uses $\lambda=0.5$ with text replay.
}
\label{tab:genppl_full}

\begin{tabularx}{\linewidth}{
@{}
>{\raggedright\arraybackslash}X
>{\centering\arraybackslash}p{0.145\linewidth}
>{\centering\arraybackslash}p{0.065\linewidth}
>{\centering\arraybackslash}p{0.085\linewidth}
r
r
r
r
@{}
}
\toprule
\textbf{Model}
& \textbf{Direction}
& \textbf{Params}
& \textbf{Data}
& \textbf{GenPPL $\downarrow$}
& \textbf{BLEU$\downarrow$}
& \textbf{Spk  $\uparrow$}
& \textbf{UTMOS $\uparrow$} \\
\midrule

\rowcolor{gray!10}
\textbf{Ground truth}
    & --
    & --
    & --
    & 43.8
    & 6.0
    & 0.74
    & 2.82 \\

\addlinespace[2pt]

\rowcolor{gray!15}
\multicolumn{8}{@{}l@{}}{
    \textbf{Speech-only baselines (no text guidance)}
} \\

Speech-only
    & S $\rightarrow$ S
    & 0.8B
    & 21.3k h
    & 151.7
    & 4.3
    & 0.42
    & 2.56 \\

Interleaved
    & S $\rightarrow$ S
    & 0.8B
    & 21.3k h
    & 164.1
    & 5.4
    & 0.37
    & 2.36 \\

\addlinespace[2pt]

\rowcolor{blue!18}
\makebox[0pt][l]{\textbf{TGSLM variants}}
    & & & & & & & \\

TGSLM
    & S $\rightarrow$ (T, S)
    & 0.8B
    & 21.3k h
    & 58.9
    & 3.9
    & 0.36
    & 2.41 \\

\quad + self-generated prompt text
    & (S, T) $\rightarrow$ (T, S)
    & 0.8B
    & 21.3k h
    & 62.2
    & 3.8
    & 0.38
    & 2.40 \\

\quad + distillation ($\lambda=0.25$)
    & S $\rightarrow$ (T, S)
    & 0.8B
    & 21.3k h
    & 56.9
    & 4.3
    & 0.36
    & 2.41 \\

\quad + distillation + text replay
    & S $\rightarrow$ (T, S)
    & 0.8B
    & 21.3k h
    & 51.0
    & 4.9
    & 0.38
    & 2.44 \\

TGSLM-Large-distill
    & S $\rightarrow$ (T, S)
    & 1.9B
    & 135k h
    & 45.0
    & 5.7
    & 0.45
    & 2.68 \\

\addlinespace[2pt]

\rowcolor{orange!18}
\makebox[0pt][l]{\textbf{Text-leading references}}
    & & & & & & & \\

Oracle prompt text
    & (T, S) $\rightarrow$ (T, S)
    & 0.8B
    & 21.3k h
    & 40.7
    & 5.4
    & 0.42
    & 2.50 \\

Whisper prompt transcript
    & (T, S) $\rightarrow$ (T, S)
    & 0.8B
    & 21.3k h
    & 57.1
    & 6.5
    & 0.41
    & 2.50 \\

\addlinespace[2pt]

\rowcolor{gray!15}
\makebox[0pt][l]{\textbf{External models}}
    & & & & & & & \\

\rowcolor{orange!8}
\multicolumn{8}{@{}l@{}}{
    \textbf{\textcolor{orange!55!black}{With limited acoustic information}}%
} \\

SpiritLM Base
    & S $\rightarrow$ S
    & 7B
    & 570k h
    & 30.4
    & 13.3
    & 0.05
    & 2.39 \\

SpiritLM Expressive
    & S $\rightarrow$ S
    & 7B
    & 570k h
    & 67.3
    & 8.7
    & 0.07
    & 2.45 \\

SLAM-Large
    & S $\rightarrow$ S
    & 1.3B
    & 81.0k h
    & 53.8
    & 10.2
    & 0.07
    & 3.46 \\

\addlinespace[1pt]

\rowcolor{green!8}
\multicolumn{8}{@{}l@{}}{
    \textbf{\textcolor{green!40!black}{With richer acoustic information}}%
} \\
~\citep{chou2026quantifying} with oracle prompt text
    & (T, S) $\rightarrow$ (T, S)
    & 0.5B
    & 10k h
    & 55.0
    & 6.6
    & 0.35
    & 2.48 \\

~\citep{chou2026quantifying} with Whisper prompt text
    & (T, S) $\rightarrow$ (T, S)
    & 0.5B
    & 10k h
    & 72.1
    & 10.7
    & 0.32
    & 2.35 \\
MiMo-Audio-7B-Base
    & S $\rightarrow$ (T, S)
    & 7B
    & \mbox{$>$100M h}
    & 14.9
    & 20.3
    & 0.40
    & 3.03 \\

Llama-Mimi-1.3B
    & S $\rightarrow$ S
    & 1.3B
    & 240k h
    & 88.7
    & 6.2
    & 0.39
    & 2.61 \\

Llama-Mimi-8B
    & S $\rightarrow$ S
    & 8B
    & 240k h
    & 53.2
    & 7.0
    & 0.41
    & 2.59 \\

\bottomrule
\end{tabularx}

\vspace{1mm}
\end{table}



\begin{table*}[t]
\centering
\small
\setlength{\tabcolsep}{4pt}
\renewcommand{\arraystretch}{1.06}

\caption{
Known-1000 evaluation.
Pass@\(K\)(\%) measures whether the answer appears in at least one of \(K\) sampled continuations.
Direction S $\rightarrow$ (T, S) denotes a speech prompt followed by a text-guided speech continuation.
\textbf{TGSLM-Large-distill} uses a distillation weight of $\lambda=0.5$ and text replay.
Post-trained models are shown in gray because they are not directly comparable to pretrained models.
}
\label{tab:known1000_full}

\begin{tabularx}{\textwidth}{
@{}
>{\raggedright\arraybackslash}X
>{\centering\arraybackslash}p{0.17\textwidth}
>{\centering\arraybackslash}p{0.08\textwidth}
>{\centering\arraybackslash}p{0.10\textwidth}
r
r
@{}
}
\toprule
\textbf{Model}
& \textbf{Direction}
& \textbf{Params}
& \textbf{Data}
& \textbf{Pass@1}
& \textbf{Pass@5} \\
\midrule

\rowcolor{gray!15}
\multicolumn{6}{@{}l@{}}{\textbf{Text-LM reference}} \\
Qwen3-0.6B backbone~\citep{yang2025qwen3}
    & T $\rightarrow$ T
    & 0.6B
    & --
    & 46.8
    & 77.9 \\
Qwen3-1.7B backbone~\citep{yang2025qwen3}
    & T $\rightarrow$ T
    & 1.7B
    & --
    & 60.8
    & 86.6 \\

\addlinespace[2pt]

 \rowcolor{blue!18}
 \multicolumn{6}{@{}l@{}}{\textbf{Our models}} \\
TGSLM-distill
    & S $\rightarrow$ (T, S)
    & 0.8B
    & 45k h
    & 9.1
    & 22.4 \\
TGSLM-Large-distill
    & S $\rightarrow$ (T, S)
    & 1.9B
    & 135k h
    & 16.8
    & 37.9 \\


\addlinespace[2pt]

\rowcolor{gray!15}
\multicolumn{6}{@{}l@{}}{\textbf{External pretrained models}} \\

\rowcolor{orange!10}
\multicolumn{6}{@{}l@{}}{
    \textbf{\textcolor{orange!35!black}{With text supervision}}
} \\

SpiritLM Base~\citep{nguyen2025spirit}
    & S $\rightarrow$ S
    & 7B
    & 570k h
    & 8.6
    & 23.5 \\

    & S $\rightarrow$ T
    &
    &
    & 15.1
    & 31.8 \\


MiMo-Audio-7B-Base~\citep{xiaomi2025mimo}
    & S $\rightarrow$ (T, S)
    & 7B
    & \mbox{$>$100M h}
    & 26.0
    & 49.5 \\

\addlinespace[1pt]

\rowcolor{green!10}
\multicolumn{6}{@{}l@{}}{
    \textbf{\textcolor{green!45!black}{Without text supervision}}
} \\

SLAM-Large~\citep{maimon2025slamming}
    & S $\rightarrow$ S
    & 1.3B
    & 81.0k h
    & 1.2
    & 5.5 \\

Llama-Mimi-1.3B~\citep{sugiura2025llama}
    & S $\rightarrow$ S
    & 1.3B
    & 240k h
    & 0.2
    & 0.6 \\

Llama-Mimi-8B~\citep{sugiura2025llama}
    & S $\rightarrow$ S
    & 8B
    & 240k h
    & 0.9
    & 2.9 \\

\addlinespace[2pt]

\rowcolor{gray!12}
\multicolumn{6}{@{}l@{}}{
    \textbf{\textcolor{black!60}{External post-trained models}}
} \\

\ptgray{Qwen2.5-Omni-3B~\citep{Qwen2.5-Omni}}
    & \ptgray{S $\rightarrow$ (T, S)}
    & \ptgray{3B}
    & \ptgray{$\sim$3.3M h$^\dagger$}
    & \ptgray{32.9}
    & \ptgray{55.1} \\

\ptgray{Qwen2.5-Omni-7B~\citep{Qwen2.5-Omni}}
    & \ptgray{S $\rightarrow$ (T, S)}
    & \ptgray{7B}
    & \ptgray{$\sim$3.3M h$^\dagger$}
    & \ptgray{44.4}
    & \ptgray{65.8} \\

\ptgray{LFM2.5-Audio~\citep{amini2025lfm2}}
    & \ptgray{S $\rightarrow$ (T, S)}
    & \ptgray{1.2B}
    & \ptgray{472k h}
    & \ptgray{35.2}
    & \ptgray{40.3} \\

\bottomrule
\end{tabularx}

\vspace{1mm}


\end{table*}

\end{document}